\documentclass[manuscript]{acmart}
\usepackage{makecell}
\usepackage{multirow}
\usepackage{tabularx,booktabs}
\usepackage{url}
\usepackage{bm}
\usepackage{hyperref}
\hypersetup{colorlinks}
\usepackage{lipsum} 
\AtBeginDocument{%
  }

\setcopyright{none}

\acmConference[Efficient LLM Algorithmic Survey]{}{Nov, 2023}{USA.}

\acmPrice{15.00}
\acmISBN{978-1-4503-XXXX-X/18/06}

\newcommand{\paratitle}[1]{\vspace{1.5ex}\noindent\textbf{#1}}
\usepackage{cleveref}

\begin{document}

\title{The Efficiency Spectrum of Large Language Models: An Algorithmic Survey}


\author{Tianyu Ding}
\affiliation{%
  \institution{Microsoft}
  \city{Hekla}
  \country{tianyuding@microsoft.com}
  }
\email{larst@affiliation.org}

\author{Tianyi Chen}
\affiliation{%
  \institution{Microsoft}
  \country{tiachen@microsoft.com}
}

\author{Haidong Zhu}
\affiliation{%
  \institution{University of Southern California}
  \country{haidongz@usc.edu}}

\author{Jiachen Jiang}
\affiliation{%
  \institution{Ohio State University}
  \country{jiang.2880@osu.edu}}

\author{Yiqi Zhong}
\affiliation{%
  \institution{University of Southern California}
  \country{yiqizhon@usc.edu}}

\author{Jinxin Zhou}
\affiliation{%
  \institution{Ohio State University}
  \country{zhou.3820@osu.edu}}

\author{Guangzhi Wang}
\affiliation{%
  \institution{Microsoft}
  \country{t-gwang@microsoft.com}}

\author{Zhihui Zhu}
\affiliation{%
  \institution{Ohio State University}
  \country{zhu.3440@osu.edu}}

\author{Ilya Zharkov}
\affiliation{%
 \institution{Microsoft}
 \country{zharkov@microsoft.com}}

\author{Luming Liang}
\affiliation{%
 \institution{Microsoft}
 \country{lulian@microsoft.com}}

\renewcommand{\shortauthors}{Ding, Chen, et al.}

\newcommand{\td}[1]{\textcolor{blue}{\textbf{#1}}}

\newcommand{\ie}{\emph{i.e.,}\xspace}
\newcommand{\aka}{\emph{a.k.a.,}\xspace}
\newcommand{\eg}{\emph{e.g.,}\xspace}
\newcommand{\wrt}{\emph{w.r.t.}\xspace}
\newcommand{\wo}{\emph{w/o}\xspace}
\newcommand{\etc}{\emph{etc}}

\newcommand{\borderline}{\td{\rule{0.25\linewidth}{0.5mm} Working until this line \rule{0.25\linewidth}{0.5mm}}}
\newcommand{\zz}[1]{{\color{blue} Zhihui: #1}}
\newcommand{\firstrefine}[1]{{\color{blue}#1}}
\begin{abstract}
\textbf{Abstract ---}  The rapid growth of Large Language Models (LLMs) has been a driving force in transforming various domains, reshaping the artificial general intelligence landscape. However, the increasing computational and memory demands of these models present substantial challenges, hindering both academic research and practical applications. To address these issues, a wide array of methods, including both algorithmic and hardware solutions, have been developed to enhance the efficiency of LLMs.  This survey delivers a comprehensive review of algorithmic advancements aimed at improving LLM efficiency\footnote{We have mostly included research works proposed prior to September 2023, along with selected contributions postdating this period. Given the extensive volume of literature in this field, it is possible that certain relevant studies may have been inadvertently missed. We welcome suggestions for additional references and are open to incorporating them in subsequent revisions.}.
Unlike other surveys that typically focus on specific areas such as training or model compression, this paper examines the multi-faceted dimensions of efficiency essential for the end-to-end algorithmic development of LLMs. Specifically, it covers various topics related to efficiency, including scaling laws, data utilization, architectural innovations, training and tuning strategies, and inference techniques. This paper aims to serve as a valuable resource for researchers and practitioners, laying the groundwork for future innovations in this critical research area. Our repository of relevant references is maintained \href{https://github.com/tding1/Efficient-LLM-Survey}{here}.
\end{abstract}




\maketitle

\section{Introduction}
\label{sec:introduction}


%
%
%
%

Large Language Models (LLMs)~\cite{zhao2023survey, shanahan2022talking, huang2022towards,chang2023survey,yang2023harnessing}, characterized by their massive scale of tens or even hundreds of billions of parameters~\cite{brown2020language,chowdhery2022palm,anil2023palm}, have become a central focus in the field of artificial intelligence. These models, exemplified by applications like ChatGPT~\cite{chatgpt2023} and Claude~\cite{claude2023}, have demonstrated impressive capabilities in a variety of general-purpose tasks, such as text summarization~\cite{yang2023exploring}, translation~\cite{hendy2023good}, question answering~\cite{robinson2022leveraging}, and even rudimentary coding~\cite{chen2021evaluating}, largely attributed to their expertise in natural language understanding. Although the exact mechanisms driving their exceptional performance remain elusive~\cite{zhu2023physics}, it is widely believed that their large size endows them with emergent abilities~\cite{wei2022emergent} not observed in smaller models, and this is seen as a key step toward achieving Artificial General Intelligence (AGI)~\cite{agi2023,bubeck2023sparks}.

\begin{figure}[t]
    \centering
    \begin{minipage}{0.32\textwidth}
           \includegraphics[width=\textwidth]{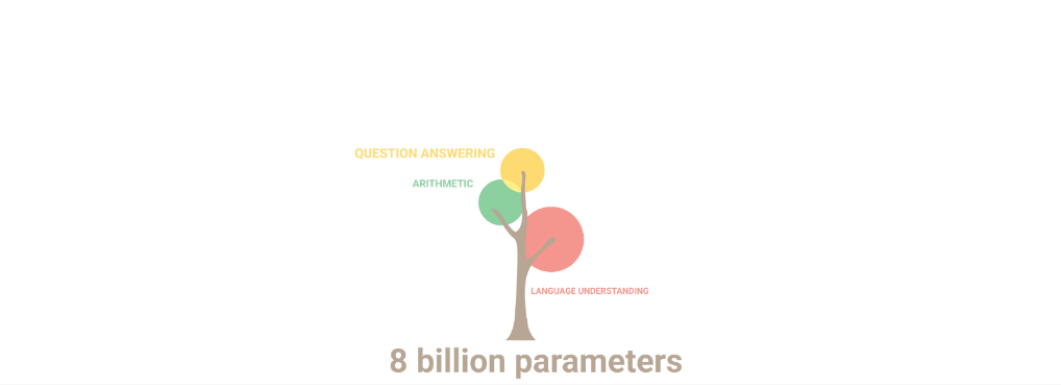}
           \captionof*{figure}{(a) 8-billion model}
        \end{minipage}
        \begin{minipage}{0.32\textwidth}
           \includegraphics[width=\textwidth]{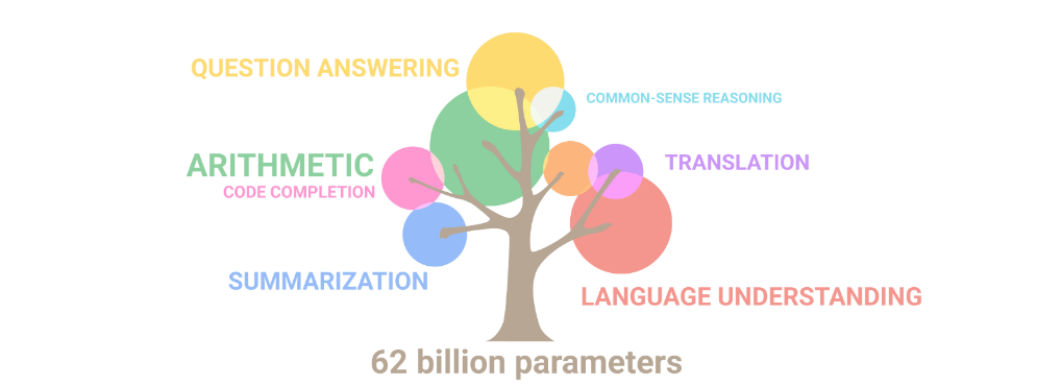}
           \captionof*{figure}{(b) 62-billion model}
    \end{minipage}
    \bigskip
        \begin{minipage}{0.32\textwidth}
           \includegraphics[width=\textwidth]{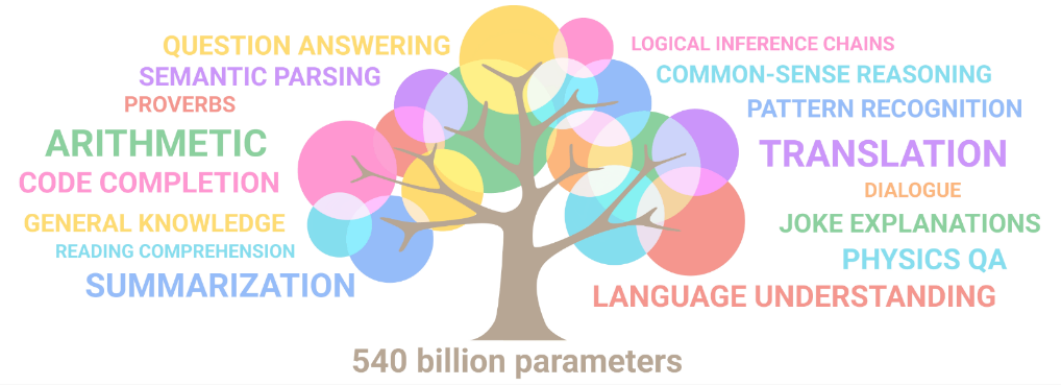}
           \captionof*{figure}{(c) 540-billion model}
    \end{minipage}
    \caption{{Capability and performance tree~\cite{palm-blog2022} of PaLM~\cite{chowdhery2022palm} across different scales (8-billion, 62-billion, 540-billion). Each circle node represents a specific capability, and its size indicates the corresponding performance level—the larger the circle, the greater the capability. As the model scale increases, performance not only improves across existing tasks but also reveals new capabilities. }}
    \label{fig:palm}
\end{figure}

{While the large size of LLMs is crucial for their capabilities (see \Cref{fig:palm}), it also presents a significant drawback: their deployment is severely limited by high computational costs and memory requirements~\cite{treviso2023efficient,zhuang2023survey,zhu2023survey,xu2023survey}. The resources needed for training these models are substantial, creating challenges in both resource allocation and model design. For example, the cost of exploring different architectures or strategies becomes prohibitive~\cite{zhao2023survey}. Furthermore, their large size makes them unsuitable for resource-constrained environments like edge devices, thus narrowing their range of applications~\cite{ahmed2020democratization}.  This computational burden also  confines the development of LLMs to large companies with abundant resources~\cite{radford2019language,brown2020language,OpenAI2023GPT4TR}. Many essential details, such as the data collection pipeline and training methodologies, remain proprietary, which hinders academic research and poses challenges for smaller companies. Additionally, the environmental impact of training these models is not to be overlooked, raising concerns about carbon emissions and ethical considerations~\cite{vinuesa2020role,wu2022sustainable,van2021sustainable}. Consequently, there is a growing emphasis on improving the \emph{efficiency} of LLMs.}

{
Motivated by this pressing need for more efficient LLMs, this survey aims to provide a comprehensive and up-to-date understanding of the subject. For the context of this paper, "efficiency" is defined as the optimization of computational and memory resources without compromising model performance. Adopting a holistic approach, we explore multiple dimensions of efficiency that are crucial for the end-to-end development of LLMs. These dimensions encompass data utilization, architectural designs, training and tuning strategies, and inference techniques, 
---essentially covering the entire pipeline of model development from algorithmic and software perspective.\footnote{Remark here that the advancements and achievements from hardware aspect are omitted in this survey.} 
While there are existing surveys that focus on specific aspects of LLM efficiency, such as data~\cite{zha2023data}, training~\cite{zhuang2023survey,zhou2023comprehensive,shen2023efficient}, tuning~\cite{zhang2023instruction}, or inference~\cite{zhu2023survey, xu2023survey},  they often do not provide a comprehensive view. Other works, like~\cite{treviso2023efficient}, have offered valuable insights into various efficiency aspects for Natural Language Processing (NLP), yet the rapidly evolving nature of the LLM field calls for an updated and comprehensive review. In contrast, our paper aims to present a more thorough and current overview of key methodologies and techniques that contribute to the development of efficient LLMs.
}

\begin{figure}[t]
    \centering
    \includegraphics[width=0.96\textwidth]{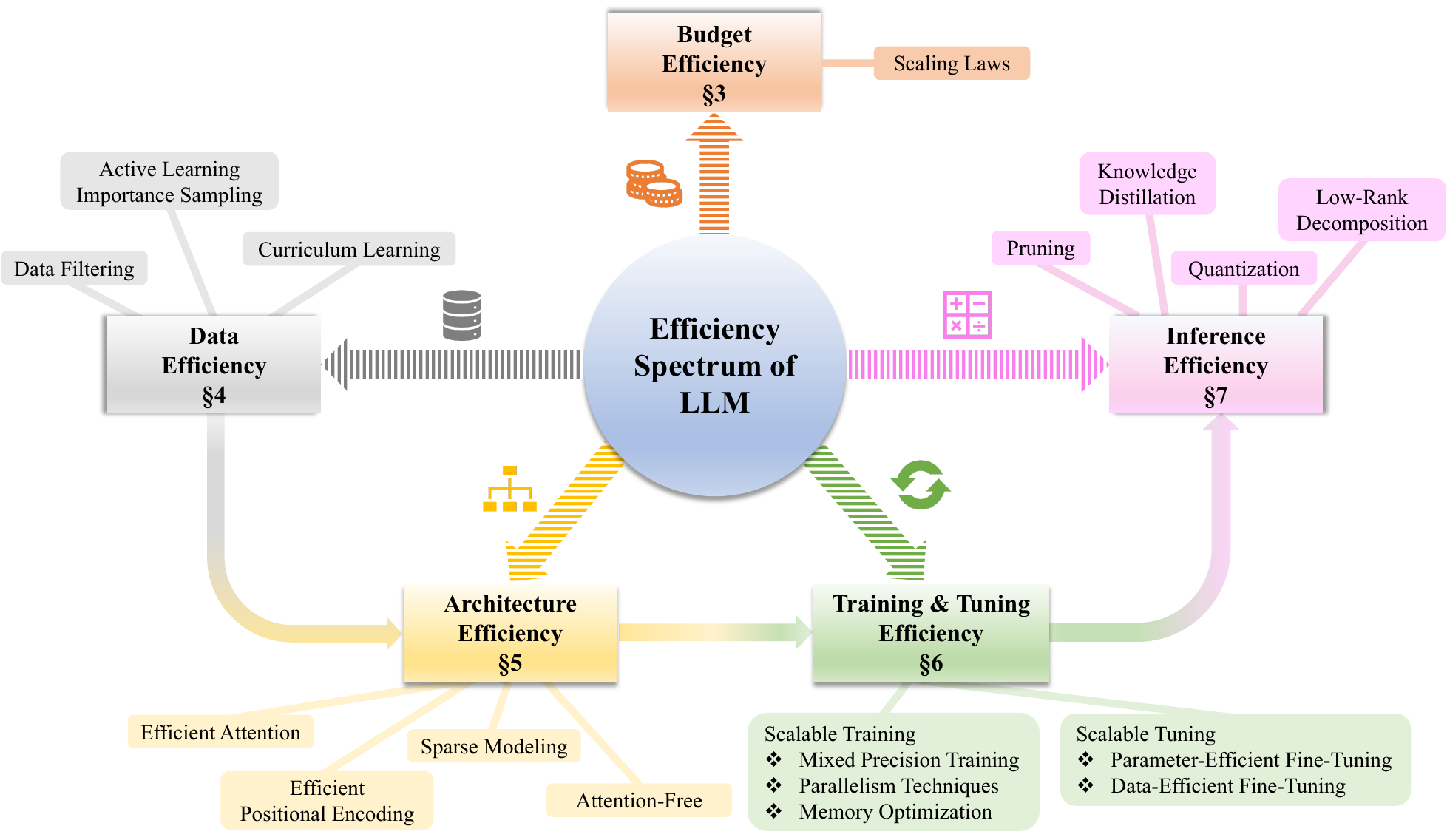}
    \caption{{The schematic overview of the multi-faceted dimensions of LLM Efficiency. This diagram illustrates the key areas covered in this survey, including data utilization, architectural designs, training and tuning strategies, and inference techniques, thereby providing a holistic view of the factors contributing to LLM efficiency.}}
    \label{fig:overview}
\end{figure}

{The remainder of this survey is organized as follows to offer a comprehensive understanding of the multiple facets of LLM efficiency from the algorithmic perspective:

$\bullet$ \Cref{sec:bg} \emph{Background} introduces the core concepts of LLMs and outlines the evaluation metrics pertinent to assessing their efficiency.

$\bullet$ \Cref{sec:budget} \emph{Budget Efficiency} examines the role of predictive approaches like scaling laws in optimizing the performance of LLMs within given resource constraints.


$\bullet$ \Cref{sec:data} \emph{Data Efficiency} focuses on techniques for optimizing data utilization, thereby reducing resource consumption without compromising performance.

$\bullet$ \Cref{sec:arch} \emph{Architecture Efficiency} reviews innovative architectural designs, providing a detailed examination of how architecture influences efficiency.

$\bullet$ \Cref{sec:train-tune} \emph{Training and Tuning Efficiency} discusses strategies for efficiently training LLMs from scratch and fine-tuning pre-trained models for specific downstream tasks.

$\bullet$ \Cref{sec:inference} \emph{Inference Efficiency} explores the realm of model compression techniques designed to accelerate inference speed and reduce memory footprint.



$\bullet$ \Cref{sec:conclusion} \emph{Conclusion}  summarizes the key findings of this survey and discusses their broader implications for the advancement of efficient LLMs.

A schematic overview of these various dimensions of LLM efficiency is presented in~\Cref{fig:overview}.
}










\section{Background}\label{sec:bg}

{
In this section, we present an overview of the core concepts that form the basis of LLMs, along with the key metrics used for assessing their efficiency.
}

\subsection{Core Concepts of LLMs}

{
Language modeling, a cornerstone in the field of NLP, aims to model the generative likelihood of word sequences and predict the probabilities of subsequent or missing tokens. This area has evolved significantly over several decades. Initially rooted in statistical language models~\cite{jelinek1998statistical,rosenfeld2000two,zhai2008statistical,bahl1989tree,chelba2013one}, the focus has gradually shifted to pre-trained neural language models~\cite{mikolov2010recurrent,mikolov2011extensions,peters2018deep,kenton2019bert,radford2019language,lewis2020bart} and, more recently, to Large Language Models~(LLMs)~\cite{zhao2023survey, shanahan2022talking, huang2022towards,chang2023survey,yang2023harnessing}. While there is no standardized definition for LLMs, they are typically distinguished by their extensive parameter sizes and extraordinary learning capabilities. In this section, we adopt the criteria outlined in~\cite{zhao2023survey}, focusing on language models with more than one billion parameters, and discuss their core concepts in detail.
}

{
\paratitle{Architectural Foundations.} LLMs can generally be categorized into two main paradigms: encoder-decoder models~\cite{peters2018deep,liu2019roberta,sanh2019distilbert,lee2020biobert,lan2020albert,raffel2020exploring}, exemplified by BERT~\cite{kenton2019bert}, and decoder-only models~\cite{radford2019language,OpenAI2023GPT4TR,brown2020language,radford2018improving,zhang2022opt,chowdhery2022palm,anil2023palm,scao2022bloom,du2022glam,rae2021scaling,hoffmann2022training,thoppilan2022lamda,touvron2023llama,touvron2023llama2,wu2023bloomberggpt}, such as the GPT series~\cite{OpenAI2023GPT4TR,radford2018improving,radford2019language,brown2020language}. BERT is trained using masked language modeling, enabling it to excel in contextual understanding by predicting masked or missing tokens. On the other hand, GPT models are trained using autoregressive modeling, which equips them with strong generative capabilities by predicting subsequent tokens in a sequence. Despite these differences, both types of models commonly rely on the Transformer~\cite{vaswani2017attention} architecture, which is particularly noteworthy for its self-attention mechanism. In self-attention, each token is represented as a \emph{key}, \emph{value}, and \emph{query}. The \emph{query} weighs the importance of other tokens, represented as \emph{keys} in understanding a particular token. These weights are applied to the \emph{values} to create context-aware representations. This mechanism allows each token in the sequence to consider all other tokens simultaneously, facilitating parallel processing of sequential data and effective capture of long-sequence dependencies. As a result, multi-head attention layers are often stacked to form deep networks in LLMs. Nowadays, decoder-only models like GPT-4~\cite{OpenAI2023GPT4TR} and LLaMa~\cite{touvron2023llama,touvron2023llama2} are becoming increasingly prevalent, yet the core architectural module of self-attention remains a constant across these variations.
}

{
\paratitle{Training Essentials.} LLMs acquire their general-purpose capabilities from an initial \emph{pre-training} phase on expansive and diverse datasets~\cite{brown2020language,yang2023harnessing}. These datasets cover a broad spectrum of sources such as books, scientific papers, code, and websites~\cite{zha2023data}. This foundational knowledge is then fine-tuned on relatively smaller datasets in a supervised manner, aimed at enabling the LLMs to adhere to human instructions, a process known as \emph{instruction tuning}~\cite{christiano2017deep,ziegler2019fine,zhang2023instruction,ouyang2022training,wang2022self,alpaca,vicuna2023,liu2023llava,zhu2023minigpt,Dai2023InstructBLIPTG,zhong2023mat,su2023pandagpt}. 
A \textit{reinforcement learning with human feedback} is then proceeded to a instructed fine-tuned LLM to further align model behavior upon human preferences and instructions~\cite{touvron2023llama2}. 
In the current landscape, decoder-only models have become the norm, particularly for their superior generative abilities. These models employ autoregressive objectives in the pre-training phase to maximize the likelihood of predicting subsequent tokens based on their preceding context. 
The scaling~\cite{kaplan2020scaling,hoffmann2022training} of this autoregressive pre-training significantly enhances LLM capabilities, as demonstrated by models like the GPT and PaLM series~\cite{anil2023palm,chowdhery2022palm}. For instance, PaLM~\cite{chowdhery2022palm} was trained on a 780B-token dataset and utilized a 540B-parameter Transformer architecture. {PaLM-2}~\cite{anil2023palm}, its successor, advances this further; its largest variant has 1.1T parameters and was trained on a more diverse, multilingual dataset with 1.5T tokens.
However, this scaling introduces its own set of challenges, requiring efficient training infrastructures and optimization strategies.  Distributed training frameworks that focus on data parallelism, such as DeepSpeed~\cite{rasley2020deepspeed} and Fully Sharded Data Parallel (FSDP)~\cite{fsdp2023}, or pipeline parallelism like Gpipe~\cite{huang2019gpipe} and PipeDream~\cite{narayanan2019pipedream}, are commonly employed. Additionally, tensor parallelism techniques like Megatron-LM~\cite{shoeybi2019megatron,narayanan2021efficient} and SARATHI~\cite{agrawal2023sarathi} are also utilized. Specialized techniques, such as mixed-precision training~\cite{scao2022bloom,nair2023int,dettmers2022gpt3} and quantization-aware training~\cite{liu2023llm,xiao2023smoothquant,lin2023awq}, are often incorporated to streamline the training process. These methods not only address the computational challenges tied to the scale of LLMs but also facilitate the development of increasingly capable models.
}

{
\paratitle{Versatile Abilities with Prompt Engineering.} One of primary mechanisms for harnessing the versatility of LLMs across diverse tasks is through prompt engineering~\cite{white2023prompt,liu2023pre,zhou2023large}. In this setting, a prompt consists of natural language instructions given by users to guide the LLM's behavior. The art of prompt engineering lies in crafting these instructions to elicit specific and contextually appropriate responses from the model. Two prominent techniques are few-shot~\cite{brown2020language,wei2022chain} and zero-shot~\cite{kojima2022large} prompting. Few-shot prompting provides the model with example tasks and corresponding solutions, while zero-shot prompting relies solely on task descriptions. This form of prompting is closely related to in-context learning~(ICL), a concept first observed in GPT-3~\cite{brown2020language}, which allows the model to adapt to new tasks without retraining. The emergent abilities of LLMs, particularly in  tackling a wide array of unseen tasks, are significantly enhanced when ICL is combined with well-designed prompts. Advanced prompting strategies like Chain-of-Thoughts~(CoT)~\cite{wei2022chain,wang2023selfconsistency}, Tree-of-Thoughts~(ToT)~\cite{yao2023tree,long2023large,xie2023decomposition}, and Graph of Thoughts~(GoT)~\cite{besta2023graph} draw inspiration from human reasoning and cognitive structures. These strategies enable LLMs to explore novel capabilities like backtracking, thought merging, and idea elimination, thereby enhancing the quality of responses in complex reasoning tasks such as arithmetic~\cite{patel-etal-2021-nlp}, commonsense reasoning~\cite{talmor-etal-2019-commonsenseqa}, and question answering~\cite{geva-etal-2021-aristotle}. As a result, prompt engineering serves as a pivotal mechanism for amplifying the versatility and effectiveness of LLMs.
}

\subsection{Evaluation Metrics for Efficiency}

{
Evaluating the efficiency of LLMs requires a multi-faceted approach that considers various performance indicators. These metrics are often presented alongside measures of accuracy and versatility to provide a holistic assessment of an LLM's overall efficiency and effectiveness. In the paragraphs that follow, we will explore key metrics commonly used to understand efficiency in the realm of LLMs.
}

{
\paratitle{Number of Parameters.} The number of parameters in an LLM is a key factor that directly affects the model's learning capacity and complexity.  These parameters, which include elements like weights and biases, are learnable during training or fine-tuning phases. A higher parameter count usually enables the model to grasp more complex data patterns, contributing to the development of various emergent abilities. However, this comes with the downside of increased computational demands for both training and inference.  Additionally, having too many parameters can lead to overfitting, especially when the training data is scarce. To mitigate this, common techniques like regularization and early stopping are frequently used.
}

{
\paratitle{Model Size.} Model size, defined as the disk space required for storing the entire model, is often the initial consideration when training a new LLM or working with a pre-trained model. Given that exceedingly large models may be infeasible to store or run, this metric is particularly crucial for real-world deployments, especially in storage-constrained environments like edge devices. Expressed in units such as gigabytes (GB) or megabytes (MB), model size is influenced by several factors. While the number of parameters plays a significant role, other elements like the data type used for parameters (\emph{e.g.}, float16, int8) and specific architectural choices also contribute. In addition to its direct impact on storage requirements, model size serves as an indirect indicator of the computational resources needed for both training and inference.
}

{
\paratitle{Floating Point Operations (FLOPs).} Floating-point operations~(FLOPs) is commonly used to gauge the computational complexity of LLMs. This metric  counts the number of floating-point operations like addition, subtraction, multiplication, and division, giving an estimate of the computation done during a single forward pass. While FLOPs offer valuable insights into computational needs  and potential energy use, they are not a complete measure. Other factors, such as system parallelism and architectural choices, also play a role in determining a model's overall computational efficiency. A higher FLOPs count usually means the model is more computationally demanding, which can be a challenge for deployment in environments with limited resources. As a result, optimizing this metric is often a key focus in the development of more efficient LLMs.
}

{
\paratitle{Inference Time / Tokens per Second.} Inference time, also known as latency or delay, measures the duration it takes for an LLM to process input and generate a response during the inference stage. Unlike FLOPs, which provide a theoretical estimate of computational needs, inference time offers a practical gauge of real-world performance. This is because it's assessed in actual deployment settings, taking into account specific hardware and optimizations. Usually expressed in milliseconds (ms) or seconds (s), this metric is crucial for real-time applications that need quick responses or have stringent latency constraints. Normalizing the inference time by time elapsed results in tokens per second, which refers to the number of tokens that a language model can process (read, analyze, generate, etc.) in one second. This is a key performance indicator that reflects the model's speed and efficiency. Achieving a balance between fast inference time / tokens per second and high generalization is a key focus in the development of efficient LLMs. 
}

{
\paratitle{Memory Footprint.} Memory footprint refers to the amount of  Random Access Memory (RAM) required to load and run a model during inference or training. This metric is crucial for understanding the model's operational demands, especially in resource-constrained environments like edge devices or servers with limited memory capacity. Expressed in MB or GB, the memory footprint includes not just the model parameters but also other runtime necessities such as intermediate variables and data structures. A larger memory footprint can limit the model's deployability and may require optimization techniques like model pruning or quantization to reduce it. 
}

{
\paratitle{Carbon Emission.} Carbon emission is an increasingly important metric in the evaluation of large models, reflecting the environmental impact of training and running these models. This metric is usually measured in terms of kilograms or tons of CO2 equivalent emitted during the model's lifecycle, from training to inference. The carbon footprint is influenced by various factors, including the energy efficiency of the hardware used, the source of electricity, and the duration of model training and operation. High carbon emissions not only have environmental implications but can also affect the social and ethical considerations of deploying LLMs. As a result, there is a growing emphasis on optimizing models to be more energy-efficient, thereby reducing their carbon footprint. This is often achieved through hardware acceleration, algorithmic improvements, or even selecting greener energy sources for data centers.
}



\section{Budget Efficiency: Scaling Laws} \label{sec:budget}

\subsection{Introduction}

{
The performance of large language models (LLMs) is significantly influenced by various factors, including training data, model size, architecture, computing resources, and the training methodology itself. Training LLMs requires extensive resources, making the conventional trial-and-error method for optimizing these factors both impractical and resource-intensive. As a result, predicting LLM performance before training is not just beneficial, but often necessary. This predictive approach allows for more effective planning and allocation of resources. For instance, consider a scenario with limited computing resources: How can we optimally balance the model size and training data to achieve minimal objective function value? Answering such questions beforehand can significantly enhance the efficiency and effectiveness of LLM training processes.
}

{
Recent research in predicting large language model (LLM) performance has concentrated on understanding the \textit{scaling law}~\cite{kaplan2020scaling}. This law delineates how LLM performance is influenced by factors such as model architecture, neural model size, computing power for training, and available data. The concept of scaling law, rooted in statistical mechanics approaches for predicting model generalization, has a rich history dating back to the early 1990s~\cite{gyorgyi1990statistical,amari1992four,seung1992statistical,barkai1993scaling}. Its relevance has been reinvigorated recently in the context of modern deep learning models~\cite{hestness2017deep,rosenfeld2019constructive,henighan2020scaling,kaplan2020scaling,hernandez2021scaling,tay2022scaling,muennighoff2023scaling,hoffmann2022training,caballero2022broken,alabdulmohsin2022revisiting,tay2021scale,sorscher2022beyond}. This section will delve into the latest advancements and insights in the scaling law as applied to LLMs, highlighting how these models evolve and perform under varying conditions.
}

\subsection{Scaling Law}

{
The work \cite{kaplan2020scaling} presents a thorough study of the empirical scaling laws of transformer-based large language models. The authors observe that model performance (objective function $L$) primarily depends on three factors: the number of model parameters $N$, dataset size $D$, and the computing budget for training. 
They demonstrate a power-law relationship between model performance (measured in objective function, $L$) and these factors. For instance, they found that the relationship between performance and dataset size can be represented as $L(D) \approx (5.4 \times 10^{13}/D)^{0.095}$. This formula suggests that as the dataset size increases, the model's performance improves following a specific pattern. While theoretical generalization bounds may suggest a similar power-law relationships, they generally do not provide specific coefficients like those identified in Kaplan et al.'s work~\cite{kaplan2020scaling}. This specificity is crucial for accurately predicting model performance. Additionally, the study highlights that transformers, known for their effective handling of long-range data dependencies, tend to outperform Long Short-Term Memory networks (LSTMs)~\cite{hochreiter1997long} as they scale. This observation underscores the potential of transformers in large-scale language processing tasks.
}

{
\paratitle{Compute-Optimal Models via Scaling Law.} When working within a fixed computational budget, it is crucial to find the right balance between model size ($N$) and dataset size ($D$). This is where the scaling law curve, $L(N,D)$, becomes a vital tool. It helps determine the most effective trade-off between these two factors. The scaling law, as observed in \cite{kaplan2020scaling}, was instrumental in designing GPT-3, a 175 billion parameter language model. Interestingly, GPT-3 was trained on fewer tokens than was typical for its time~\cite{brown2020language}. Different forms of the scaling law curve have led to the development of diverse models, as seen in subsequent studies~\cite{hoffmann2022training,caballero2022broken,alabdulmohsin2022revisiting}. A notable application of these predicted scaling laws is found in \cite{hoffmann2022training}. Their research revealed that many previously trained LLMs, including Gopher \cite{rae2021scaling}, could have achieved better performance within the same compute budget. They demonstrated this by training a smaller model, Chinchilla, with 70 billion parameters, which outperformed the larger 280 billion parameter Gopher model \cite{rae2021scaling} while using a similar compute budget.
}

{
\paratitle{Scaling Law for Transfer Learning.} 
While the scaling behavior of pretrained LLMs has been extensively studied and exhibits clear predictability, it becomes less clear when predicting the performance of pretrained LLMs on downstream tasks.
The work in \cite{hernandez2021scaling} investigates the scaling behavior when fine-tuning a pretrained model and demonstrates that favorable scaling laws, akin to those in \cite{kaplan2020scaling}, apply to transfer and few-shot settings in NLP tasks. In comparison to models trained from scratch, pretrained models exhibit a more advantageous scaling law in low-data scenarios. Different scaling behaviors between upstream and downstream configurations are observed in \cite{tay2021scale}, indicating that, in addition to model size, the model's architecture plays a critical role in downstream fine-tuning. Specifically, the study demonstrates that redesigned models can attain similar downstream fine-tuning quality while having 50\% fewer parameters and training 40\% faster when compared to the widely adopted T5-base model~\cite{2020t5}.
}

{
\paratitle{Scaling Law in the Data-Constrained Regime.} What if the training data are limited? The work \cite{muennighoff2023scaling} examines a data-constrained regime and observes that, for a fixed compute budget, training with up to 4 epochs of repeated data yields negligible changes in the objective function compared to a single epoch\footnote{Unlike general machine learning models, which are typically trained with multiple epochs, it is common practice to train Language Models (LLMs) using a couple of epochs \cite{komatsuzaki2019one, brown2020language}.}. However, further increasing the number of epochs will lead to a decrease in the performance of the learned model.
}

{
\paratitle{Effect of Data Quality.} A pivotal question in the realm of machine learning is whether the quality of data can lead to a transition from power-law to exponential scaling in model performance. The work by~\cite{sorscher2022beyond} provides an intriguing insight into this matter. They demonstrate that for certain vision classification tasks, the objective function can exhibit exponential scaling with an increase in dataset size, deviating from the traditional power-law scaling observed with pruned datasets. While this phenomenon is initially observed in vision tasks, recent research, including works by \cite{eldan2023tinystories,li2023textbooks,gunasekar2023textbooks}, expands this concept to other domains. These studies explore the impact of high-quality data in tasks like generating coherent English, coding, and common sense reasoning. They suggest that high-quality data can significantly alter the scaling laws' trajectory. This change indicates the potential for more efficient models, which, despite being trained on fewer data tokens yet with high quality, could match the performance of large-scale models trained on vast datasets without sufficient quality constraints. This shift in understanding the role of data quality could revolutionize the approach to training and optimizing LLMs.
}

{
\paratitle{Effect of Architecture.} In the domain of model scaling, conventional wisdom, as supported by studies like \cite{kaplan2020scaling,henighan2020scaling}, suggests that the inherent attributes of models, such as the width or depth of Transformers, have a minimal impact on performance. However, the work by~\cite{tay2022scaling} presents a contrasting viewpoint. This study delves into the influence of different architectural designs on the scaling law and reveals that architecture plays a crucial role in scaling processes. Tay et al.~\cite{tay2022scaling} demonstrate that depending on the scale, the most effective model architecture can vary significantly. This finding complements previous assumptions and underscores the importance of considering architectural variations in the quest for optimal model performance at different scales.
}

\section{Data Efficiency}\label{sec:data}

\subsection{Introduction}

{
The insatiable demand for data by large-scale models has significantly fueled the growth of the data collection and preparation industry. However, this reliance on vast datasets, often accumulated over years, introduces substantial challenges in model training. These include not only prolonged training durations but also escalated costs due to extensive power consumption and the need for larger data storage capacities. Consequently, finding ways to use data more efficiently in both training and validation phases is of paramount importance. In this section, we will delve into strategies and considerations for enhancing data efficiency, addressing how to maximize the utility of data while mitigating the associated costs and resource demands.
}


\subsection{Data Filtering}

{
Data filtering is pivotal in directing training focus towards more informative samples, thereby eliminating irregular characters or patterns, rather than concentrating on examples with lesser informational value. 

\paratitle{Deduplication.} A prime data filter is \textit{removing duplications}, \textit{i.e.}, \textit{deduplication}. This straightforward yet efficacious approach not only abbreviates training duration but also enhances model performance, as evidenced by \cite{lee2021deduplicating}. The utility of this de-duplication operation is evident at both the pre-training and fine-tuning stages of model development. In both pre-training and fine-tuning, the researchers utilizes techniques such as MinhashLSH~\cite{leskovec2014mining},  CC-NET~\cite{wenzek2020ccnet}, and adversarial filtering~\cite{zellers2018swag}, as demonstrated by~\cite{mishra2020we,zhang2022opt,bowman2015large}, for purging duplicates from the training datasets. 
}

{
\paratitle{Data Undersampling.} Beside deduplication, \textit{data undersampling}, also referred to as instance selection, emerges as another promising data filtering technique \cite{zha2023data}. This approach aims to reduce the volume of training samples by sub-sampling large datasets, yet crucially retains the distribution characteristics of the original training data. Moreover, this sub-sampling process can help mitigate issues of data imbalance. For instance, Prusa et al. \cite{prusa2015using} demonstrated the effectiveness of random undersampling in majority classes, which serves dual purposes: it reduces redundancy in the training set and balances the data distribution across various classes. Additionally, MESA~\cite{liu2020mesa} represents a novel advancement in this area by adopting Meta Learning to learn how to undersample vast datasets effectively. 
These techniques highlight the strategic importance of selective data use, not only in terms of quantity but also in ensuring quality and balance. 
}

\subsection{Active learning / Importance Sampling}\label{sec.important_sample}

{
Active Learning or importance sampling helps a machine learning algorithm achieve better or equivalent performance with fewer annotated training samples. Their applications in training with extensive datasets predates the emergence of LLMs~\cite{katharopoulos2017biased,settles2012al}. These methodologies strategically reduce the total number of training samples by applying various criteria. They aim to optimize the data collection and selection procedure by selecting and annotating only the most useful instances during the data annotation procedure~\cite{settles2009active,settles2011theories,shen2023efficient}. The essence lies in the ability to prioritize samples based on their significance to the learning process, thereby optimizing the training efficiency for models dealing with large-scale data. 
}




{
\paratitle{Gradient Norm.} 
Katharopoulos and Fleuret \cite{katharopoulos2018not} introduced an approach based on the upper bound of the gradient norm, along with proposing an estimator for variance reduction through importance sampling.  Furthermore, Zhang et al. \cite{zhang2019autoassist} implemented a selector mechanism to identify samples with larger gradients within a batch, treating them as valid approximations for importance sampling. According to their methodology, samples with smaller gradients are deemed `good points' in a given epoch, necessitating increased focus on `bad points', \textit{i.e.},  samples with larger gradients. Upon these paradigm, training and fine-tuning over the samples with larger gradients could enhance the model performance more effectively.
}

{
\paratitle{Objective Function / Predicted Score.}  Beyond the gradient, Katharopoulos and Fleuret also suggested in a separate work \cite{katharopoulos2017biased} that the objective function (loss value) itself could serve as a viable metric for importance sampling. Expanding on this concept, Jiang et al. \cite{jiang2019accelerating} introduced `selective back-propagation'. This method accelerates training by omitting the back-propagation stage for training samples exhibiting low loss, a strategy claimed to enhance convergence and outpace traditional Stochastic Gradient Descent (SGD) on full datasets. In the context of text retrieval, a similar \textit{relevance sampling}~\cite{salton1990improving,gale1992method} has been proposed. It leverages the scores predicted by current model to evaluate the relevance of the samples and let the annotators only annotate those with higher scores. Such methods may focus on the samples with high scores thereby resulting in overfitting. Some researchers turn to \textit{uncertainty-based sampling} methods \cite{lewis1995sequential,tang2002active,gal2017deep,siddhant2018deep,wang2022uncertainty}. This thread of methods defines that the instances with lower uncertainties are more useful for improving performance and are worthier of annotation.
}

{
\paratitle{Diversity Sampling.} In contrast to the sampling methods which find difficult examples to train and annotate, methods that follow the strategy of \textit{diversity sampling} \cite{xu2003representative,hu2010off,bodo2011active,sener2017active,gissin2019discriminative,su2022selective,zhang2022allsh} want to enhance the heterogeneity of training data. For diversity sampling, there are two major approaches: iterative selection and cluster-based selection. Methods that belong to iterative selection iteratively examine if each  instance can help improve the training data diversity and only annotate qualified instances. For example, Vote-k proposed in \cite{su2022selective} adopts the cosine similarly between the embedding of each unlabeled instance and its k-nearest neighbors to define instances representativeness; ALLSH~\cite{zhang2022allsh} represents the diversity according to the data instance's local sensitivity~\cite{chapelle2009semi}. Cluster-based methods will cluster unannotated datasets and select instances based on the cluster labels. Methods in this category have explored many clustering strategies for instance samplings, such as (weighted) K-means~\cite{zhdanov2019diverse,chen2023maybe,maekawa2022low,yu2022actune}.
}

{
\paratitle{Hybrid Sampling.} Gradient or objective-function based sampling necessitates a relatively reliable initial Large Language Model (LLM), as sample selection depends on the model's predicted outputs, a process often termed a \textit{warm start} \cite{yuan2020cold}. On the other hand, diversity sampling can accommodate a \textit{cold start} scenario. However, this approach may inadvertently introduce outliers, potentially compromising the final model's performance.
To address the trade-off between both sampling threads, recent researches have shifted towards \textit{hybrid sampling methods} that integrate both elements to surmount the challenge \cite{kirsch2019batchbald,ash2019deep,yuan2020cold,margatina2021active,chung2023increasing,siddiqui2022metadata}. For instance, Yuan et al.  \cite{yuan2020cold} incorporated BADGE \cite{ash2019deep} into language processing. This method involves transforming unannotated instances into representations that encapsulate model confidence, followed by clustering these transformed instances. In parallel, Margatina et al. \cite{margatina2021active} introduced Contrastive Active Learning (CAL). CAL selectively acquires contrastive samples from the pool of raw unannotated data, defining them as instances that are proximate in the model's feature space (\textit{e.g.}, sharing similar vocabulary or model encodings) but yield divergent predictive likelihoods. Experimental results affirm that CAL realizes a more effective balance compared to individual strategies, showcasing its potential in refining sampling processes.
}

{
\paratitle{Other Sampling.} In addition to the above techniques, there exists a spectrum of sampling methods that operate on a broader scale. Chen et al. \cite{chen2022mask} introduced a novel approach by masking parameters to halt the back-propagation from certain inputs. This method preserves the forward graph, ensuring that the statistical characteristics of all batches are maintained, especially in layers such as normalization, thereby reflecting the overall dataset's mapping. Taking a different approach, Xie et al. \cite{xie2023data} implement importance sampling in a reduced feature space. This strategy facilitates the tractability of importance weight estimation across the expansive text space, leveraging Kullback-Leibler (KL) reduction to efficiently train Large Language Models. Moreover, Sujit et al.  \cite{sujit2022prioritizing} advocate for clipping training samples based on their relevance in reinforcement learning contexts, prioritizing samples according to their importance. Yao et al.  \cite{yao2022nlp} propose a keyword-based retrieval system for selecting relevant training samples in semi-supervised language fine-tuning, usually picking the top-k neighbors for each target example during training. These diverse methodologies highlight the evolving landscape of importance sampling, extending its application beyond standard batch-centric approaches to encompass broader training strategies for LLMs.
}

\subsection{Curriculum Learning}
Curriculum learning \cite{elman1993learning,bengio2009curriculum} is a strategy that aims to improve the model training efficiency by carefully designing the feeding order of the instances in the training data. The principle of this approach is to initiate training with simpler samples or subtasks and progressively escalate to more challenging ones. Two critical components are integral to the design of 
a curriculum learning method. \textit{(i)} The \textit{difficulty metric} (or difficulty criterion), responsible for ranking the training samples based on their complexity. This metric serves as a guide to categorize training instances from the simplest to the most complex. \textit{(ii)} The \textit{pacing function} (also known as curriculum scheduler or arrangement), which determines the rate at which these ranked samples are fed to the model training. This function modulates the learning curve, ensuring that the model is not overwhelmed by the complexity of the tasks too early.

\paratitle{Difficulty Metric. } In the realm of curriculum learning for natural language processing, the most widely used difficulty metric is perhaps the sequence length \cite{li2021curriculum,li2022stability,geiping2023cramming}. The underlying assumption is that processing longer sentences poses greater challenges than shorter ones. Another prevalent metric is vocabulary rarity \cite{wang2022language,zhao2020reinforced}, based on the intuition that sentences with less frequently used words in the training set are inherently more complex to comprehend. In addition, this metric could be measured by the uncertainty sampling principle in active learning, where the uncertainty indicated by other pre-trained models could serves as a gauge of difficulty as well.

For fine-tuning in specific downstream tasks, researchers have innovated task-specific difficulty metrics. A notable example is in paraphrase generation, where Kadotani et al. \cite{kadotani2021edit} proposed using the edit distance between paraphrased sentence pairs as a metric, approximating the extent of required transformations. These custom metrics, tailored to the nuances of specific tasks like paraphrase generation, often outperform general metrics such as sentence length or word rarity. In the context of neural machine translation (NMT), defining sample difficulty is more complex. In addition to sentence length and word rarity, some studies \cite{kocmi2017curriculum} have incorporated linguistically-motivated features. For instance, the number of coordinating conjunctions in a target sentence has been used to estimate sentence complexity. Such linguistically-informed metrics provide a more nuanced understanding of difficulty in tasks where traditional metrics may fall short, reflecting the evolving landscape of curriculum learning in NLP.



\paratitle{Pacing Function.} In the curriculum learning, the pacing function plays a crucial role in dictating the progression of training complexity. A common approach involves utilizing predefined step-wise functions, such as linear, root, or exponential curves. Typically, this process starts by defining the total training steps $T$, the highest difficulty level $d_{max}$ and the lowest difficulty level $d_{min}$. The difficulty level for each training step $t$ is then determined by the chosen curve. Apart from step-wise methods, stage-wise pacing methods offer an alternative. An example of this is Shortformer \cite{press2020shortformer}, which employs a two-stage approach. In the first stage, it exclusively trains on short sequences, while in the second stage, it focuses solely on longer sequences, with a fixed length of 3072 tokens in the Shortformer model.

Beyond the separate determination of difficulty metrics and pacing functions, self-paced learning \cite{kumar2010self} presents a more integrated strategy. This approach involves simultaneously selecting easier samples and learning a new parameter vector in each iteration. The number of samples chosen is regulated by a weight that gradually increases, eventually encompassing the entire training dataset. This method has found applications in fields like machine translation \cite{wan2020self} and dialogue generation \cite{zhu2021combining}, showcasing its versatility and effectiveness in various NLP tasks.
Curriculum learning has been successfully adopted to improve the data efficiency of many LLM downstream pretraining and finetuning procedures. However, how to choose the difficulty metric and pacing speed is non-trivial and requires experimental investigation for certain tasks or models.





\section{Architecture Efficiency}\label{sec:arch}






\subsection{Introduction}

Recently, the Transformer family~\cite{vaswani2017attention} has been the dominant architecture for language modeling, owing to its strong parallelism over recurrent methods such as RNNs~\cite{mikolov2010recurrent}. However, its substantial computational cost renders the overall architecture inefficient in processing and handling long inputs. In particular, one of the key operations in the Transformer architecture is the attention mechanism. It typically requires quadratic complexity with respect to sequence length for computation and is thus significantly slow when processing long text inputs~\cite{hassid2022much}. Reducing the computation required by the attention operation~\cite{tay2022efficient} becomes a direct solution to improve the architecture's efficiency, benefiting both the training and inference stages. Toward this end, 
researchers are exploring solutions for more \textit{efficient attention}~\cite{child2019generating, dao2022flashattention, dao2023flashattention} along with different types of \textit{positional encoding}~\cite{press2021train, chi2022kerple, chi2023dissecting, li2023functional, su2021roformer, peng2023yarn, ruoss2023randomized, kazemnejad2023impact}, or leveraging the inherent sparsity within the model to avoid activating all parameters during the feedforward computation 
with \textit{sparse modeling}~\cite{du2022glam, shen2023flan}. Additionally, some recent works have directly replaced the attention mechanism with alternative architectures, introducing \textit{attention-free} methods~\cite{peng2023rwkv, dao2022hungry, poli2023hyena, sun2023retentive} to the fold. In this section, we cover these four primary directions with the latest progress.


\subsection{Efficient Attention}

The Transformer model, as introduced by Vaswani et al.~\cite{vaswani2017attention}, utilizes a vanilla attention mechanism that computes dense pairwise relations in the input sequence. This results in quadratic complexity. However, recognizing that not all these relations hold equal significance, recent research has focused on methods to streamline this process. These methods aim to identify and maintain only the most crucial relations, thereby enhancing the efficiency of the attention calculation~\cite{child2019generating,dao2022flashattention,dao2023flashattention,liu2023blockwise,liu2023ring,ainslie2023gqa,zhuang2022long,ding2023longnet}. In this subsection, we include the discussion of two main branches: \textit{(i)} the use of fast or sparse attentions~\cite{child2019generating,kitaev2020reformer,choromanski2020rethinking,dai2019transformer,martins2021infty,ding2023longnet,zhuang2023survey,ma2022mega} and \textit{(ii)} IO-aware attention calculation with hardware co-design~\cite{dao2022flashattention,dao2023flashattention,ham2021elsa,ham20203,tao2023flashdecoding,hong2023flashdecoding++}.
Both approaches reduce hardware loading time for efficient attention calculations. 

\paratitle{Fast Attention Calculation.} In the realm of fast attention, researchers are developing innovative strategies to enhance efficiency. A primary focus is on \textit{attention factorization}, which aims to reduce attention calculations that are often unnecessary in certain contexts. This technique is particularly useful when dealing with lengthy sequential inputs, where direct pairwise attention computations become computationally intensive. By employing attention factorization, computational demands can be significantly reduced, transforming 2-D computations into more manageable 1-D formats~\cite{child2019generating, zaheer2020big, ainslie2020etc, ma2022mega}. Furthermore, these factorized attention methods are designed to discern and emphasize the attention differences between closely positioned tokens and their respective changes over time. This nuanced approach ensures that the computational resources are dedicated to the most impactful elements of the data. Another innovative method involves the use of \textit{frequency-based techniques}, such as Fast Fourier Transform (FFT) and hash representations. These techniques model attention in a manner that aligns well with hardware capabilities, making them more efficient for practical applications~\cite{zhuang2022long, dao2019learning_butterfly, sun2021sparse}. They filter out near-zero attentions and focus computational efforts on the most significant ones for the final calculations. Such selective attention ensures that resources are not wasted on processing relatively unimportant data, further optimizing the overall efficiency of the model.

Moving away from directly calculating pairwise attention, some methods~\cite{dao2022monarch, fu2023monarch, ding2023longnet, liu2023blockwise} explore the possibility of computing attention at a \textit{block level}, which enables parallelization of the computation, significantly enhancing efficiency. For instance, the Monarch framework~\cite{dao2022monarch} and its advanced version, Monarch Mixer (M2)~\cite{fu2023monarch}, adopt a novel strategy. They sparsify the dense attention matrix by decomposing it into a combination of permutation and block diagonal matrices. This decomposition allows for more efficient processing of attention calculations. Furthermore, the Blockwise Self-Attention (BST) method~\cite{liu2023blockwise,liu2023ring} introduces blockwise computation for both self-attention and feed-forward networks. This technique aims to lower memory requirements typically associated with traditional attention mechanisms. Moreover, some methods like LongNet~\cite{ding2023longnet} incorporate dilated attention in place of the original dense attention, allowing the processing of much longer token sequences, thereby expanding the capabilities of LLMs.

\paratitle{Hardware-Related Efficient Attention.} Along with designing more efficient attention mechanisms at the software level, a significant focus has shifted to optimizing these mechanisms at a hardware level. One of the main challenges in this domain is efficiently utilizing computational resources, such as High Bandwidth Memory (HBM) and Static Random-Access Memory (SRAM), on GPUs. In this regard, recent advancements like FlashAttention~\cite{dao2022flashattention} and its successor, FlashAttention-2~\cite{dao2023flashattention}, have emerged. FlashAttention reconsiders the attention computation by adopting an I/O-centric perspective. It minimizes data transfers between HBM and SRAM, addressing a critical bottleneck in GPU processing. This method integrates block-wise softmax value calculations and updates with statistics. Such integration eliminates the conventional requirement of computing the softmax function only after all attention values have been determined. Building upon FlashAttention~\cite{dao2022flashattention}, FlashAttention-2~\cite{dao2023flashattention} further optimizes work partitioning and reduces non-matrix multiplications, capitalizing on the GPUs' optimization for matrix operations. These algorithms are tailored to hardware considerations and accelerate the models on GPU machines. {Based on FlashAttention, FlashDecoding~\cite{tao2023flashdecoding} and FlashDecoding++~\cite{hong2023flashdecoding++} split the keys/values in smaller chunks for parallelization of partial attention, and FlashDecoding++~\cite{hong2023flashdecoding++} further introduce asynchronized softmax with unified max value and flat GEMM optimization with double buffering for further speed up, along with the heuristic dataflow for adaption on hardware resources.}

In the quest to enhance Large Language Model (LLM) systems, some researchers are creatively drawing inspiration from current hardware architectures. A notable example is the PagedAttention~\cite{kwon2023vllm}, which adapts virtual memory and paging techniques, commonly used in operating systems, to overcome memory limitations in LLMs. PagedAttention introduces an innovative approach to memory management by emulating the virtual memory system. It segments the Key-Value (KV) cache associated with a request into blocks, as opposed to relying on pre-allocated, contiguous memory. This method significantly reduces memory fragmentation, a common issue in traditional LLM memory allocation strategies. As a result, it allows the LLM to process longer sequences within the constraints of limited memory resources.

\subsection{Efficient Positional Encoding}

Since LLMs may need to process long sequences as input, the absolute positional encoding (APE) used in the vanilla Transformer~\cite{vaswani2017attention} falls short of this requirement. To enhance the architecture's efficiency, researchers are exploring novel positional encoding (PE) methods that can accommodate longer sequences with relative positions~\cite{press2021train, chi2022kerple, chi2023dissecting, li2023functional} or rotary positional encoding~\cite{su2021roformer, peng2023yarn}. They are also seeking more generalizable solutions through randomized positional encoding~\cite{ruoss2023randomized} or even omitting positional encoding~\cite{kazemnejad2023impact}. We discuss some of the latest developments in this section.

\paratitle{Addition-based Relative Positional Encoding.} Relative positional encoding methods
utilize the relative position between two tokens rather than the absolute position of a single token. Some of them encode the relative positions and add the encoded positions to the subsequent attention, referring to \textit{addition-based} relative positional encoding methods. T5~\cite{raffel2020exploring}, TISA~\cite{wennberg2021case}, and FIRE~\cite{li2023functional} are representatives of this paradigm. In these models, the position embedding is applied to the interaction between the query and key elements within the self-attention mechanism, a departure from the earlier focus on the absolute position of individual tokens.
The relative positional encoding in T5~\cite{raffel2020exploring} translates the relative position difference into a scalar bias value using a lookup table and employs the same embedding for all out-of-distribution (OOD) sequence lengths. TISA incorporates~\cite{wennberg2021case} a trainable Gaussian kernel that focuses on the positional differences between tokens. 
FIRE~\cite{li2023functional}, on the other hand, employs progressive interpolation with a normalized position index by dividing the index difference between tokens by the smaller of the two indices. Compared to APE, relative positional encoding (RPE) offers a more effective way of modeling the relative distances between tokens. This not only enhances the model's understanding of token relationships but also facilitates length extrapolation, a critical feature for handling varied and complex sequences in language processing.


\paratitle{Relative Positional Encoding with Decay Functions.} Another trend is to employ trainable relative positional encodings (RPE) that use decaying functions. This approach, exemplified by models like ALiBi~\cite{press2021train}, KERPLE~\cite{chi2022kerple}, and Sandwich~\cite{chi2023dissecting}, aims to focus the model’s attention predominantly on neighboring tokens. The use of decaying functions in these methods ensures that the attention diminishes as the distance between tokens increases. ALiBi introduces a linear decaying function to model the relationship between tokens, particularly effective for capturing the diminishing relevance of tokens as their distance increases. KERPLE~\cite{chi2022kerple} uses two variations of conditionally positive definite (CPD) kernels: a logarithmic variant and a power variant. These sophisticated kernels decay the connection between two tokens during RPE computation to adaptively model the decreasing significance of distant token relationships. Sandwich~\cite{chi2023dissecting}, meanwhile, adopts a series of cosine functions to represent the differences between tokens. Sandwich leverages the periodic nature of cosine functions to capture the cyclical patterns in token relationships. By diminishing attention between distant positions, these methods ensure the model's focus remaining on the more immediate and contextually relevant tokens rather than the tokens that are far away.

\paratitle{Rotary Positional Encoding.} Beyond addition-based relative positional encoding, which adds encoded positions to the attention calculation, there are RPE methods that utilize rotary matrices for position embeddings~\cite{su2021roformer,chen2023extending,peng2023yarn}. 
RoPE~\cite{su2021roformer} introduces two rotation matrices to rotate the query and key vectors. The rotation angle is proportional to their absolute positions, which is then integrated into the dot product attention mechanism. This manner allows RoPE to generate attention based on the relative distance between tokens, instead of directly computing their relative differences. However, RoPE faces limitations in generalizing to sequence lengths beyond what it was trained on. Building upon RoPE, PI~\cite{chen2023extending}  extends its capabilities with Position Interpolation (PI). After fine-tuning on a moderate amount of data, PI shows a promising ability to handle very long context windows, addressing one of RoPE's primary limitations. YaRN~\cite{peng2023yarn} further advances this field by introducing NTK-aware interpolation and dynamic NTK interpolation. This method effectively addresses the loss of high-frequency information in scenarios with and without fine-tuning on limited datasets. YaRN’s approach significantly improves the model's ability to expand context size without the necessity for extensive fine-tuning. The common thread among these methods is their use of rotary matrices in the query and key vectors, a technique that has shown promising results in establishing more effective RPE in Transformer models.


\paratitle{Other Positional Encodings.} Exploring beyond relative positional encoding (RPE) methods, Randomized PE~\cite{ruoss2023randomized} and NoPE~\cite{kazemnejad2023impact} present approaches that do not rely on modeling the consecutive positions of tokens in the input query. Intriguingly, they posit that by including positions outside the length of the training distribution or by forgoing positional encodings altogether, the model can handle out-of-distribution cases with longer token lengths and exhibit enhanced generalizability on downstream tasks.
Randomized PE~\cite{ruoss2023randomized} employs a number greater than the longest sequence encountered during training. It randomly samples from a range of integers, using them as indices after sorting. This approach enables the model to generalize to longer sequences during inference, though it requires prior knowledge of the maximum token length. On the other hand, NoPE completely forgoes the positional encoder in the self-attention mechanism. It demonstrates that the model's self-attention can inherently learn the RPE across tokens in a sentence. This omission not only simplifies the model architecture but also shows promising results in terms of generalizability, especially for sentences with queries extending beyond the training distribution.


\subsection{Sparse Modeling}

In the quest to optimize Transformers for efficiency, another key area of research focuses on integrating sparse modeling within these attention-based architectures. This approach is pivotal in reducing computational demands, especially in models with a large number of parameters. Two primary directions have emerged in sparse modeling: the Mixture of Experts (MoE)~\cite{du2022glam, chen2023sparse, shazeer2017outrageously, lepikhin2020gshard, zoph2022designing, rajbhandari2022deepspeed, mustafa2022multimodal} and Sparsefinder~\cite{treviso2021predicting} from different manners. 


The MoE approach~\cite{chen2023sparse, shen2023flan, fedus2022switch, du2022glam, yi2023edgemoe, chen2023pipeline}, incorporates multiple branches or `experts' in the model, each specializing in different subtasks. During inference, only a subset of these paths is activated, maintaining computational efficiency while potentially enhancing performance. This design enables models like GLaM to scale impressively, activating only 99 billion parameters during inference despite having over 1.2 trillion parameters in total. Further developments in MoE, such as Sparse MoE~\cite{chen2023sparse}, address issues like representation collapse, ensuring more equal activation of experts and efficient information processing. On the other hand, Sparsefinder~\cite{treviso2021predicting} takes a different approach by focusing on uncovering sparsity within the attention mechanism itself. This method identifies key patterns through the attention scheme, which helps in efficiently allocating computational resources to the most impactful areas of the model.

\subsection{Attention-free}

One significant drawback of the vanilla attention mechanism~\cite{vaswani2017attention} is the quadratic complexity of attention computation, making it especially inefficient for handling long sequences. Although efficient / sparse attention offers some relief, its worst-case theoretical complexity remains unchanged. To address this, various attention-free methods have been proposed, providing alternatives that avoid the computation of the quadratic attention matrix~\cite{peng2023rwkv, dao2022hungry, poli2023hyena, sun2023retentive, zhai2021attention, gu2021efficiently}. These methods could be largely categorized into those that replace the attention mechanism with recurrent computation~\cite{zhai2021attention, peng2023rwkv, sun2023retentive}, and those that discretize state space representations~\cite{gupta2022diagonal, gu2021efficiently, gu2022parameterization, mehta2022long,gu2023mamba,smith2022simplified}. Notably, these new methods like RWKV~\cite{peng2023rwkv}, H3~\cite{dao2022hungry}, Hyena~\cite{poli2023hyena}, RetNet~\cite{sun2023retentive} and Mamba~\cite{gu2023mamba} achieve performance comparable to the standard Transformer. RWKV~\cite{peng2023rwkv} utilizes recurrent neural networks (RNNs) to streamline sequence processing, thereby reducing the complexity of handling long sequences. H3~\cite{dao2022hungry}, based on state space models (SSMs), offers an efficient alternative for data representation and processing. Hyena~\cite{poli2023hyena} presents itself as a drop-in replacement for traditional attention mechanisms, simplifying the Transformer architecture. RetNet~\cite{sun2023retentive} introduces a multi-scale retention module coupled with a feed-forward network module, enhancing parallelism and recurrence, which significantly improves efficiency in both training and inference phases. {Mamba~\cite{gu2023mamba} includes the selection operation based on state space models for as compression and further improve the efficiency with hardware-related optimization.} We include the complexity analysis of these methods compared with the vanilla Transformer with an input query of length $n$ in Table~\ref{tab:arch_complexity}. It provides an overview of how each method scales in complexity, offering insights into the advancements in the attention-free technologies.

\begin{table}[t]
\caption{Comparisons of time and memory cost in inference evaluation between Transformer and attention-free methods when using a sequence of length $n$ as input. Models with more $+$ under performance are of better performance {of the perplexity numbers on the in-domain validation set and other out-of-domain corpora.}} 
\label{tab:arch_complexity}
\centering
\def\litem{1.3}
\def\lmethod{2}
\def\lsep{0.06}
{
\begin{tabular}{p{\litem cm}p{\lsep cm}p{\lmethod cm}<{\centering}p{\lmethod cm}<{\centering}p{\lmethod cm}<{\centering}} 
\toprule
Method && Time Cost &  Memory Cost & Performance  \\
\midrule
Transformer~\cite{vaswani2017attention} &&  $O(n)$    & $O(n^2)$      & +++(+)\\
RWKV~\cite{peng2023rwkv}                &&  $O(1)$    & $O(n)$        & +\\
H3~\cite{dao2022hungry}                 &&  $O(1)$    & $O(n\log n)$  & ++\\
Hyena~\cite{poli2023hyena}              &&  $O(n)$    & $O(n\log n)$  & +\\
RetNet~\cite{sun2023retentive}          &&  $O(1)$    & $O(n)$        & +++\\
{Mamba~\cite{gu2023mamba}}                &&  $O(1)$    & $O(n)$        & +++\\
\bottomrule
\end{tabular}
}
\vspace{-5mm}
\end{table}

\section{Training and Tuning Efficiency}\label{sec:train-tune}

\subsection{Introduction}

{The development of training and tuning techniques for LLMs must address the challenges posed by the ever-increasing size of data and models. This section delves into the efficiency aspects crucial for both scalable training and tuning of LLMs, highlighting key areas of focus.}

{\paratitle{Memory Efficiency.} The rapid growth in the number of parameters in large transformer models, increasing by approximately $410\times$ every two years, presents significant memory challenges. This growth has outpaced the expansion of GPU memory, which has seen only a $5\times$ increase (from 16GB to 80GB) over the same period. The actual memory consumption during training far exceeds the raw parameter count, encompassing model states (parameters, gradients, optimizer states), as well as residual states (intermediate activations, temporary buffers, memory fragmentation). Given these constraints, single GPU setups are insufficient for handling entire models, necessitating distributed training approaches like tensor parallelism (TP) and pipeline parallelism (PP) for effective memory management.}

{\paratitle{Computation Efficiency.} While distributed training offers potential benefits in speeding up the training of large models, it also introduces complexities that affect scalability. A notable observation is the decrease in FLOPs per GPU when training is distributed across multiple GPUs, as opposed to a single GPU setup. This decrease stems from the challenges in efficiently utilizing an increasing number of computational resources. Therefore, scalability becomes a crucial element in boosting computation efficiency during the training process, particularly in multi-GPU settings.}

{\paratitle{Communication Efficiency.} This aspect relates to the exchange of parameters and gradients between different devices or layers during training. Techniques like all-reduce are employed to synchronize gradients across all devices at the end of backward propagation in data parallel training. The goal is to minimize the volume of communication data during collective operations such as broadcast, reduce, all-reduce, and all-gather.}

{In short, training and tuning LLMs is a complex challenge that demands a comprehensive approach. An integrated strategy that considers all these efficiency aspects is vital for the effective and scalable training and tuning of LLMs. The subsequent sections will provide a detailed exploration of these aspects.}

\subsection{Scalable Training}



\subsubsection{Stable Training Strategies}

{During the pre-training of LLMs, ensuring training stability is a critical aspect of efficiency. Training instability, often manifested as vanishing or exploding gradients, can significantly hinder the training process. To mitigate these issues, careful selection and adjustment of hyperparameters are essential. One effective approach involves the strategic manipulation of batch size. For example, models like PaLM~\cite{chowdhery2022palm} gradually increase their batch size from 1 million to 4 million tokens during training. This gradual scaling helps in accommodating the model's growing capacity to process larger data volumes without compromising stability. Another key hyperparameter is the learning rate, where the warm-up cosine scheduler is commonly employed. This scheduler initially increases the learning rate during the early stages of training (typically 0.1\% to 0.5\% of total training steps) and then implements a cosine decay strategy. This approach gradually reduces the learning rate to about 10\% of its peak value, ensuring a balance between rapid learning and stability as training progresses. The choice of optimizer also plays a pivotal role in stabilizing the training of LLMs. Optimizers like Adam~\cite{kingma2014adam} and AdamW~\cite{loshchilov2017decoupled} are popular choices for models such as GPT-3~\cite{brown2020language} and OPT~\cite{zhang2022opt}, owing to their momentum feature that accelerates convergence by leveraging past gradient information. Additionally, the Adafactor~\cite{shazeer2018adafactor} optimizer, known for its GPU memory efficiency, is utilized in models like PaLM and T5~\cite{raffel2020exploring}. Beyond hyperparameter tuning, implementing stabilizing strategies like weight decay and gradient clipping is common to prevent exploding gradients. However, even with these measures, training loss spikes can still occur, often influenced by both the current model state and the data being processed. To address this, models like PaLM and OPT employ a strategy of restarting the training from a previous checkpoint when a spike is detected, effectively skipping over the data that triggered the instability. This approach ensures not only training stability but also efficient use of computational resources by avoiding prolonged periods of unproductive training.}

\subsubsection{Mixed Precision Training}

{In the realm of LLM pre-training, mixed precision training emerges as a critical strategy for enhancing both memory and computational efficiency. Traditionally, neural network training involves storing weights, gradients, and activations in full-precision (FP32) format. However, for extremely large models, this approach can be resource-intensive. To address this, reduced-precision formats like FP16 or INT8 are adopted. These formats not only reduce memory usage but also expedite communication processes within the model. In addition, modern GPUs are typically more adept at handling FP16 computations compared to FP32, offering a further boost in computational speed.

Despite these advantages, transitioning directly from FP32 to FP16 can sometimes lead to performance degradation~\cite{hubara2017quantized, zhou2016dorefa} due to issues like overflow or underflow inherent in FP16. To circumvent these challenges, the automatic mixed-precision (AMP)~\cite{micikevicius2017mixed} method has been developed. AMP maintains a master copy of weights in FP32 while employing FP16 for computations during the forward and backward passes. Post-calculation, the weights are converted back to FP32 for updating the master weights. This method, coupled with a loss scaling technique that preserves small gradient values, enables AMP to match the accuracy of FP32 training without the need for extensive hyperparameter tuning. Further advancements in precision reduction have led to the introduction of Brain Floating Point (BF16)~\cite{kalamkar2019study}, a novel half-precision format. BF16, designed to cover the same range as FP32 by allocating more bits to the exponent and fewer to the significand compared to FP16, has demonstrated state-of-the-art performance and more reliability than mixed precision under FP16. 

Another innovative approach is Activation Compressed Training (ACT)~\cite{chen2021actnn}, which focuses on compressing activations to an average of 2 bits across multiple tasks. ACT computes gradients using a compressed version of activations saved during the backward process, leading to a significant reduction in memory requirements for activations. This compression enables training with substantially larger batch sizes, ranging from $6.6\times$ to $14\times$ larger than traditional methods. Overall, mixed precision training stands as a testament to the evolving landscape of LLM training, where efficiency and performance are continually balanced through innovative techniques.}

\subsubsection{Parallelism-Based Techniques}

{
Parallelism in the training of LLMs is a strategy that involves distributing the computational workload across multiple accelerators, such as GPUs or TPUs. This approach is crucial for managing the substantial data and complex computations required in LLM training, facilitating the development of more advanced and capable models. In this section, various parallel training schemas are discussed.
}

{
\paratitle{Data Parallelism (DP).} Data parallelism~\cite{li2014scaling,xing2015petuum,li2020pytorch,fan2021dapple} a straightforward yet effective form of distributed training. In this approach, the dataset is divided into smaller subsets, which are then processed in parallel across multiple accelerators. The model is replicated across these devices, with each replica operating on a separate unit. Each unit then independently performs forward and backward computations on its assigned subsets. A key aspect of DP is the synchronization of gradients at the end of each training step. The gradients calculated on each device are averaged, resulting in a gradient representative of the entire batch. This process ensures that despite the parallel processing, the model learns consistently across all subsets of data. DP is particularly noted for its ability to maximize GPU utilization. However, it requires high-bandwidth interconnects to efficiently handle the communication demands between devices. By leveraging DP, training large-scale LLMs becomes more feasible, enabling faster development cycles and the exploration of more complex model architectures.
}

\begin{table}[t]
\caption{Summary of different parallelism strategies for efficiency.} 
\label{tab:parallelism}
\begin{tabular}{ccccc}
\toprule
\multicolumn{2}{c}{\multirow{2}{*}{Parallelism Strategy}}                                                                          & \multicolumn{3}{c}{Resource Efficiency}                  \\ \cmidrule{3-5} 
\multicolumn{2}{c}{}                                                                                                            & Memory    & Computation       & Communication                 \\ \midrule
\multicolumn{2}{c}{Data Parallelism (DP)}   &  Low     &   High   &   High  \\ \midrule
\multirow{4}{*}{Model Parallelism (MP)} & \multirow{2}{*}{\begin{tabular}[c]{@{}c@{}}Tensor Parallelism (TP)\\ (Intra-layer)\end{tabular}}   & \multirow{2}{*}{High} & \multirow{2}{*}{Low} & \multirow{2}{*}{Low} \\
  &   &                   &                   &                   \\ \cmidrule{2-5} 
& \multirow{2}{*}{\begin{tabular}[c]{@{}c@{}}Pipeline Parallelism (PP) \\ (Inter-layer)\end{tabular}} & \multirow{2}{*}{High} & \multirow{2}{*}{Low} & \multirow{2}{*}{High} \\
 &   &       &       &                   \\ \bottomrule
\end{tabular}
\end{table}

{
\paratitle{Model Parallelism (MP).} Model Parallelism is an alternative approach to DP, focusing on dividing the model itself across multiple accelerators. This method is particularly useful for handling models with a large number of parameters and sizes, especially when a single GPU lacks the capacity to store the entire model. Model Parallelism can be further categorized into two types: Tensor Parallelism (TP) and Pipeline Parallelism (PP).

$\bullet$ Tensor Parallelism~\cite{shoeybi2019megatron,shazeer2018mesh,xu2021gspmd} is a form of \emph{intra-layer} model parallelism. It involves dividing the tensors of individual layers across multiple accelerators, allowing for the training of models that are larger than the memory capacity of a single GPU. A notable example of this approach is Megatron-LM~\cite{shoeybi2019megatron}, which provides strategies for slicing different components of transformer parameters, such as MLP layers, self-attention layers, and output embedding layers. This slicing can be done either horizontally or vertically. Initially, TP focused on splitting two-dimensional matrices, but it has since evolved to include multi-dimensional splitting. Frameworks like Colossal-AI~\cite{xu2021efficient, wang2022tesseract, bian2021maximizing} have expanded TP to higher dimensions and introduced sequence parallelism~\cite{li2022sequence,korthikanti2023reducing} for handling sequence data. While TP is efficient in terms of memory usage, it requires high interconnect bandwidth for effective layer communications.

$\bullet$ Pipeline Parallelism~\cite{jia2019beyond,huang2019gpipe,narayanan2019pipedream,narayanan2021memory,narayanan2021efficient,yang2021pipemare,kosson2021pipelined,kim2023bpipe}, on the other hand, is a form of \emph{inter-layer} model parallelism. It involves splitting the layers of a model across multiple accelerators in a pipeline configuration. Each accelerator is responsible for computing a different layer and then passing its output to the next, akin to an assembly line. This setup allows for sequential yet concurrent processing of both forward and backward passes. While one segment of the model processes one part of the data, other segments can work on different parts simultaneously. The key to the efficiency of PP lies in its ability to keep all parts of the model active and productive, though it requires careful scheduling to minimize idle time on the GPUs.

Gpipe~\cite{huang2019gpipe}, one of the earliest proposals for PP, combines PP with mini-batch splitting. It segments a large model across multiple GPUs and processes input mini-batches as smaller micro-batches. This approach allows for the efficient training of significantly large models. GPipe also uses a strategy called rematerialization~\cite{chen2016training} to reduce memory usage by recalculating activations during backward propagation instead of storing them. However, GPipe still faces memory inefficiencies due to the need to cache activations during backward computations. PipeDream~\cite{harlap2018pipedream,narayanan2019pipedream} further refines the PP approach with its One Forward pass followed by One Backward pass (1F1B) strategy. This method allows for the immediate backward propagation of a micro-batch following its forward pass, enabling earlier stages of the pipeline to start their backward computations sooner. PipeDream also employs asynchronous gradient updates using different versions of weights and optimizes memory allocation across the pipeline stages for more efficient processing. BPipe~\cite{kim2023bpipe} introduces a technique to balance memory usage throughout the pipeline. It leverages idle memory in later stages to support earlier stages, significantly speeding up the training process for large models like GPT-3 $96$B and GPT-3 $134$B by $1.25$ to $2.17$ times compared to Megatron-LM. TeraPipe~\cite{literapipe} addresses the challenge of training large models with extensive sequence lengths, which can lead to smaller mini-batches and increased idle time in the pipeline. In transformer architectures, some layers' calculations don't depend on future hidden states. TeraPipe uses this fact to enable parallel processing by dividing the input sequence. It uses dynamic programming to effectively split the sequence at the best points across tokens, improving the efficiency of parallel processing.
}

{
\paratitle{Automated Parallism.} Automated Parallelism has become a key approach in scaling up modern LLMs, combining various parallelism methods for optimal performance. Systems like DeepSpeed~\cite{rasley2020deepspeed}, Megatron-LM~\cite{shoeybi2019megatron}, and Colossal-AI~\cite{xu2021efficient} have adopted a $3$D parallelism approach, which involves distributing training data uniformly across workers, manually partitioning the model, and distributing layers in each pipeline stage. However, this manual orchestration of parallelism types is complex and not easily adaptable across different models and computing environments.

To streamline this process, automated parallelization solutions are being developed. These solutions aim to speed up model deployment and ensure adaptability across different models. As models and computing clusters grow, the complexity of parallelism configurations also increases. Tofu~\cite{wangsupporting} tackles this challenge with a dynamic programming algorithm that optimizes dataflow graph partitioning. Dapple~\cite{fan2021dapple} focuses on minimizing pipeline latency through optimal partitioning strategies. However, these solutions are currently limited to combining data parallelism with only one model parallelism type, due to the complex interactions between different parallelism strategies. Alpa~\cite{zheng2022alpa} takes a more comprehensive approach by organizing data, model, and pipeline parallelism into a hierarchical structure. It uses integer linear programming for model parallelism plans and dynamic programming for pipeline parallelism plans. Alpa is comparable to specialized systems like Megatron-LM in training GPT models, demonstrating its effectiveness in handling complex parallelization challenges.

FlexFlow~\cite{jia2019beyond} extends the concept of 3D Parallelism by proposing a method to divide operation output tensors across different dimensions (Sample, Operation, Attribute, Parameter). Each operation in the computation graph is assigned a specific parallelization configuration. To find the best parallelization strategy, FlexFlow uses an execution simulator that predicts the time required to run an operator graph on a given device topology. It then employs Markov Chain Monte Carlo sampling to systematically search for the optimal strategy, considering both the operator graph and device topology. 
}

\subsubsection{Memory Optimization}

{In the realm of training LLMs with increasing sizes, the memory needed to store model parameters, gradients, and optimization states grows significantly. This challenge is particularly acute in DP, where each GPU traditionally stores a complete copy of the model's parameters, leading to considerable memory redundancy.  Efficient memory optimization strategies are essential to train larger models on limited hardware resources, balancing the memory load across different components of the training infrastructure.

ZeRO~\cite{rajbhandari2020zero} addresses the issue of memory redundancy in data parallelism by partitioning the memory load across GPUs. Instead of each GPU storing the entire set of model parameters, gradients, and optimizer states, ZeRO divides these elements, allowing each GPU to hold only a portion of the data. The remaining data can be retrieved from other GPUs as needed. This approach includes three key strategies: parameter partitioning, gradient partitioning, and optimizer state partitioning, each targeting a specific aspect of the model's memory requirements. Building on ZeRO, ZeRO offload~\cite{ren2021zerooffload} extends these concepts to enable the training with the usage of both CPU and GPU capabilities, offloading some computations and storage to the CPU to alleviate the memory burden on GPUs. However, this offloading increases communication between the CPU and GPU, which can become a bottleneck if not managed carefully. The strategy involves viewing the training process as a data flow graph, with different computation nodes assigned to different devices. The forward and backward processes are handled by the GPU, while parameter updates and precision conversions are managed by the CPU. This approach aims to minimize CPU computation and reduce communication overhead, ensuring efficient use of CPU and GPU resources in the training process. 

Integrating these advancements, systems like DeepSpeed~\cite{rasley2020deepspeed} offer different levels of memory optimization. The first stage is ZeRO-DP (Data Parallelism) which optimizes memory by partitioning only the optimizer states across GPUs. The second stage is ZeRO-R (Reduction and Partitioning) to further reduce memory usage by partitioning gradients and optimizer states. The third stage is ZeRO-Infinity that extends the memory optimization beyond what is available on the GPU, utilizing both CPU and NVMe memory to enable training of extremely large models. 
}

\subsection{Scalable Tuning}

{
Large Language Models trained on massive and varied datasets have demonstrated remarkable general problem-solving capabilities. However, their performance can be significantly enhanced for specific domains or tasks through targeted adaptation.  
In recent years, a range of techniques has emerged to facilitate this adaptation process. This section discusses two primary approaches for the efficient adaptation of pretrained LLMs: \textit{(i)} parameter-efficient fine-tuning, which involves incorporating adapter layers or fine-tuning existing parameters of the pretrained models, and \textit{(ii)} the integration of task-specific context via prompt engineering. These methods represent key strategies in tailoring LLMs to specific applications, ensuring both their versatility and effectiveness in diverse NLP tasks.
}

\subsubsection{Parameter-Efficient Fine-Tuning (PEFT)}

The substantial size of pretrained LLMs makes them expensive or impractical to fully fine-tune the entire models for the downstream task or application domain. To avoid directly fine-tuning the full LLMs, a range of parameter-efficient tuning methods have a variety of parameter-efficient tuning methods have emerged. These methods focus on refining LLMs by adjusting or introducing a small number of trainable parameters, while keeping most or all of the original pretrained parameters fixed. Such methods typically attain commendable performance and bring significant reductions in the quantity of trainable parameters. They enhance both memory and computational efficiency in comparison to full parameter tuning, offering more practical solutions for adapting LLMs to specific tasks.

\paratitle{Partial Parameter Tuning.} A straightforward yet effective approach in adapting LLMs is partial parameter tuning, where only a selected fraction of pretrained parameters are fine-tuned, leaving the rest unchanged. This method has been widely demonstrated. For example, the works~\cite{kovaleva2019revealing,lee2019would} fine-tune only a few final layers, achieving up to 90\% of the performance of a fully fine-tuned model. Xie et al.~\cite{xie2022hidden} involves selecting a subset of layers for fine-tuning based on the variability in their hidden states, particularly for classification tasks. Additionally, BitFit \cite{zaken2022bitfit} presents an alternative strategy by adjusting only the bias terms in transformer-based LLMs, yielding competitive performance. These examples underscore the potential of partial parameter tuning as a resource-efficient way to adapt LLMs for various applications. However, these methods typically lack detailed principle to guide how to select a subset of parameters for further tuning.

\paratitle{Model-Adapter tuning.} To tackle the issue of selecting specific parameters for fine-tuning, the technique of adapter tuning has been introduced, which involves augmenting the pre-trained model with additional small-scale learnable blocks, known as adapters~\cite{houlsby2019parameter}. Such approaches maintain the integrity of the pre-trained model, yet embed adapter blocks into one or several modules of the pretrained LLMs. These adaptors typically take the form of compact bottleneck layers. One example is comprising a two-layer MLP (Multi-Layer Perceptron) with a nonlinearity function and a small number of neurons in the hidden layer. The adaptor integration can be executed in series~\cite{houlsby2019parameter} or in parallel~\cite{pfeiffer2020mad} with the attention and feed-forward layers of the Transformer architecture, or outside of the Transformer architecture ~\cite{peng2023towards}. To further enhance the reuse and versatility of adapters, AdapterHub~\cite{pfeiffer2020adapterhub} has been developed. This framework allows for the dynamic integration of pre-trained adapters, catering to a variety of tasks and LLMs. Although the use of adapters accelerate the fine-tuning process and mitigates storage requirements, its does modify the computational graph by adding depth or width to each transformer layer. Such modification results in a slight increase in inference latency, as observed in studies~\cite{ruckle2021adapterdrop}, where inference speeds were found to be slower by approximately 4-6\%.

\paratitle{Parameter-Adapter tuning.} Another related approach is to directly add an adapter to the model parameters. Denoting the pre-trained network parameters as $\bm{\theta}$, this class of techniques expands the model parameters to $\bm{\theta} + \Delta \theta$, with $\theta$ being fixed and $\Delta\theta$ being learned by low-rank approximations.  
An implementation of this technique is diff-pruning~\cite{guo2021parameter} that learns task-specific sparse parameters $\Delta\theta$ by adding a sparse promoting regularization during fine-tuning. The method LoRA \cite{hu2021lora} learns low-rank transformations for each linear layer. In particular, LoRA reparameterize the weight matrix as $\theta+\Delta \theta\approx \theta + BA$, where the pretrained weight matrix $\theta$ is fixed, yet the low-rank matrices $B$ and $A$ are learnable. 
In LoRA, all weight matrices share a constant intrinsic rank for each low-rank sub-matrix, while not accounting for the varying importance across different modules. AdaLoRA~\cite{zhang2023adaptive} addresses this limitation by dynamically allocating parameter budgets to weight matrices based on their importance scores. It assigns higher ranks to more critical incremental matrices, capturing more detailed task-specific information, while reducing the rank of less important matrices to avoid overfitting and save computational resources. SoRA\cite{ding2023sparse} introduces an optimize-able gate that dynamically adjusts the rank of the incremental matrix using the proximal gradient method. In QLoRA~\cite{dettmers2023qlora}, the pretrained model is initially quantized to 4 bits. Subsequently, a small set of learnable low-rank adapter weights are augmented and fine-tuned using backpropagated gradients over the quantized weights. QLoRA can match the performance of 16-bit full-parameter fine-tuning even with 16-bit, 8-bit, or 4-bit adapters.

\subsubsection{Data-Efficient Tuning.} Data-efficient tuning refers to the process of updating a limited set of prompt parameters for downstream tasks, instead of fine-tuning the pretrained LLM. It is typically achieved through prompt tuning, where the weights of a pretrained model are kept fixed, yet only the added prompt tokens are adjusted. Such approaches enable a more efficient use of data and often yield enhanced performance, particularly as the scale of the model parameters increases.

\paratitle{Prompt Tuning.} Prompt tuning is a technique used to enhance the performance of LLMs in supervised downstream tasks. It formulates the downstream task into a masked language problem and converts the original token input into a template and masking certain tokens unfilled for the LLMs to complete. By modifying the tunable template embedding, prompt tuning aims to improving performance in the downstream tasks via reducing the distribution shift between the pretrained tasks and the specified downstream tasks. This method also enables the LLM to engage in few-shot or even zero-shot learning, especially useful in scenarios with limited supervised data, by generating new prompt templates.

Traditional methods required manual design of prompt templates and verbalizers, which often resulted in sensitive and varying efficacy. However, recent advancements in prompt learning have led to the automation and optimization of prompt construction. AutoPrompt~\cite{shin2020autoprompt} introduces a gradient-based approach to automate the search for effective templates. LM-BFF~\cite{gao2020making} offers a more efficient solution for automated prompt generation by searching for label words and using a T5-based template generation method in the discrete prompt space. To tackle the challenges of discrete optimization, Prefix-Tuning~\cite{li2021prefix} recommends parameterized prompts, where only the prompt is fine-tuned while the LLM remains unaltered. P-tuning~\cite{liu2023gpt} breaks away from the conventional constraint that templates must be composed of natural language, transforming template construction into a continuous parameter optimization challenge. CP-tuning~\cite{xu2023making} advocates the use of contrastive learning to automatically learn the distribution of embeddings, serving as a substitute for manual verbalizer design. UPT~\cite{wang2022towards} introduces the Prompt-Options-Verbalizer paradigm, facilitating joint prompt learning across diverse NLP tasks and encouraging LLMs to acquire task-invariant prompting knowledge.

\section{Inference Efficiency}\label{sec:inference}

\subsection{Introduction}

{
The enormous number of parameters in Large Language Models (LLMs) poses significant challenges for deployment on cloud services and resource-limited devices, leading to high maintenance costs for inference support. Consequently, accelerating inference has become a pressing issue garnering attention from both industry and academia. One common way is to construct compact model that could reach competitive performance to the full model, which methods can be broadly classified into four categories: pruning, knowledge distillation, quantization, and low-rank decomposition. Pruning techniques focus on identifying and eliminating redundancy within the operators of Deep Neural Networks (DNNs), thereby creating more streamlined versions. Knowledge distillation involves transferring insights from larger, more complex 'teacher' models to smaller, more efficient 'student' models, helping to maintain high performance even in reduced models. Quantization reduces computational load and storage requirements by representing floating-point numbers in LLMs with fewer bits. Low-rank decomposition approximates the heavy weight matrices in LLMs through low-rank matrices, further economizing computational resources. It's worth noting that some of these approaches require specialized computing libraries and hardware to achieve realistic resource savings and speed-up.}



\subsection{Pruning} Pruning techniques aim at identifying the redundancy inside operators of LLMs. Existing pruning techniques can be broadly classified as unstructured, semi-structured, and structured pruning. 

\paratitle{Unstructured Pruning.} Unstructured pruning yields fine-grained sparsity wherein zero elements are randomly distributed across the trainable parameters~\cite{jaiswal2023emergence,sauceoutlier2023,shao2023one,zhang2023dynamic,kurtic2023sparse,xia2023flash,sun2023simple,chen2020neural,chen2021orthant,ding2021cdfi,ding2022sparsity}. These unstructured pruning methods show that LLMs can be pruned to at least 50\% sparsity in one-shot, with(out) retraining, at minimal loss of accuracy. Meanwhile, larger LLMs are more compressible in the manner of unstructured pruning. Although unstructured pruning could bring theoretical inference speedup, the speedup is not easily reflected in reality due to the contiguity issue during sparse operations. Accelerating a DNN with high fine-grained sparsity typically requires the supports of specialized designed software and hardware. There exist computing libraries such as FSCNN~\cite{ji2022fscnn} which could outperform standard DNN runtime under sufficiently high unstructured sparsity, yet largely not extended to transformer architectures. 

\paratitle{Semi-Structured Pruning.} To mitigate the issue of unstructured pruning, semi-structured pruning is recently proposed where N:M sparsity is exampler~\cite{zhou2021learning}. N:M sparsity stays between unstructured pruning and structured pruning that every $M$ contiguous elements contain exactly $N$ non-zero elements. Nvidia~\cite{choquette2021nvidia} introduced the Ampere Tensor Core GPU architecture (e.g. A100 GPUs) and proposed the 2:4 fine-grained structured sparsity scheme that enables Sparse Neural Network to be accelerated on this hardware at inference time. This scheme places a constraint on the allowed sparsity pattern: For every contiguous array of four weights, two are pruned, yielding a 50\%-sparse net. The resulting regular structure of the weight matrix allows one to compress it efficiently and to reduce memory storage and bandwidth by operating on the nonzero weights only. Importantly, Nvidia currently considered exclusively the 2:4 ratio; other are not accelerated yet.

\paratitle{Structured Pruning.} Structured pruning removes entire neurons, channels, or other meaningful structures, thus preserving the functionality of remaining DNN that are amenable to efficient computation~\cite{chen2021only,chen2020half}. Prior structured pruning methods require manual interventions to figure out removal structures, which is inconvenient. Recent works such as OTO~\cite{chen2023otov2,chen2021only,chen2023towards} and torch-pruning~\cite{fang2023depgraph} propose dependency graph analysis to automatically find out the structures that could be removed. However, their application onto LLMs is facing significant challenges, due to the requirements of massive computational resources and the unavailable training datasets of both pretraining and instructed fine-tuning datasets~\cite{brown2020language}.  Consequently, the paradigms of structured pruning on LLMs could be largely categorized as pruning under \textit{limited} or \textit{full} resources.  For the limited-resource setup, LLM-Pruner~\cite{ma2023llm} provides modular importance score calculator to rank removal structures. The importance score is typically computed upon full gradient calculation which may be resource-consuming if with full models. A rapid post-training phase with limited instructed fine-tuning data is followed to recover lost knowledge to some extent. LoRAPrune \cite{zhang2023pruning} uses Low-Rank-Adaptor (LoRA)~\cite{hu2021lora} during the pruning  stage to reduce the resource requirements, yet still face significant performance degradation to the full LLMs. To recover and preserve knowledge more effectively, LoRAShear~\cite{chen2023lorashear} is recently proposed in the limited resource setup. LoRAShear utilizes a novel structure sparse optimizer called LoRA Half-Space Projected Gradient (LHSPG) to conduct progressive structured pruning and transfer the knowledge. Unlike the prior works only using instructed-fine-tuning data, a multi-stage knowledge recovery mechanism is applied for LoRAShear to effectively narrow down the performance gap between the full and compressed LLMs. For full-resource setups, Sheared-LLaMA ~\cite{xia2023sheared} performs structured pruning on original LLMs to create compact models that outperform equally sized LLMs trained from scratch. However, it requires significant GPU power and data resources, which may not be feasible for the public users. Compared with the relatively mature domain of structurally pruning middle-small size DNNs, structured pruning on LLMs is still in the early stage and awaiting for further explorations.



\subsection{Knowledge Distillation}

The concept of knowledge distillation involves utilizing supervisory signals from a large, more capable `teacher' model to train a compact `student' model. This approach often results in the student model surpassing the performance of a similarly sized model that was trained without such guidance~\cite{hinton2015distilling}. Knowledge distillation can be largely categorized into response-based, feature-based, and relation-based knowledge distillation~\cite{gou2021knowledge}. Response-based knowledge focuses on the final output layer of the teacher model. The hypothesis is that the student model will learn to mimic the predictions of the teacher model~\cite{hinton2015distilling,ba2014deep}. A trained teacher model also captures feature-based knowledge of the data in its intermediate layers, which is especially pertinent for deep neural networks~\cite{romero2014fitnets,zagoruyko2016paying}. Knowledge that captures the relationship between feature maps can also be used to train a student model, referring as relation-based~\cite{yim2017gift}. Initial research in NLP domain primarily concentrated on the distillation of task-specific models~\cite{kim2016sequence}. Later on, more studies have shifted their focus towards distilling pre-trained models, which can subsequently be fine-tuned for specialized downstream tasks~\cite{sanh2019distilbert,liu2020fastbert,jiao2019tinybert}. Recently, there emerge distillation methods for LLMs~\cite{zhang2023lifting,li2023symbolic,hsieh2023distilling,wang2023scott,wu2023lamini,chen2023disco,anand2023gpt4all,gu2023knowledge,jiang2023lion,chen2023mcc,sahu2023promptmix}. One main focus of the current knowledge distillation methods on LLMs lies in how to generate and utilize challenging (instructed) samples~\cite{wu2023lamini,chen2023disco,jiang2023lion} to more effectively transfer the knowledge from teacher to student model, which has some overlapping to the data undersampling methods in Section~\ref{sec:data}. Chain-of-thought prompting is commonly used in distillation approaches~\cite{li2023symbolic,hsieh2023distilling} to accomplish the data generations. {However, current knowledge distillation methods for LLMs suffers from the absence of a standardized objective function, which is typically task specific and requires tuning efforts. Furthermore, the recent strategy of employing student-generated outputs to mitigate the discrepancies between training and inference has led to a substantial increase in computational costs. Addressing these challenges, DistiLLM~\cite{ko2024distillm} has been recently introduced, consisting of two key components. \textit{(i)} A skew Kullback-Leibler (KL) divergence loss with moderate skew value, which offers stable gradients and lower approximation errors compared to other alternatives, including the generalized Jensen–Shannon divergence. \textit{(ii)} An adaptive off-policy technique, designed to improve the efficiency and effectiveness of using student-generated outputs.}

\subsection{Quantization}

{
Quantization methods can be divided based on the necessity for retraining~\cite{gholami2022survey}. Quantization-Aware Training (QAT) mandates model retraining, adjusting its weights to recover accuracy post-quantization~\cite{bai2020binarybert,kim2021bert,shen2020q,zhang2020ternarybert}. In contrast, Post-Training Quantization (PTQ) achieves quantization without any retraining~\cite{cai2020zeroq,li2023q,oh2022non,yao2022zeroquant,yuan2023rptq}. Although QAT often yields superior accuracy, it is frequently impractical for Large Language Models (LLMs) due to the prohibitive costs of retraining and, usually, the lack of access to original training data and processing infrastructure. As a result, most research on LLM quantization gravitates towards PTQ techniques.

From another perspective, quantization methods can be broadly classified into uniform and non-uniform approaches~\cite{gholami2022survey}. Uniform quantization, as explored in works like SPQR~\cite{dettmers2023spqr}, GPTQ~\cite{frantar2022gptq}, and others~\cite{huang2023output,kim2021bert}, involves dividing the range of weights into equally sized bins. This method has become popular for its ability to accelerate computation by allowing arithmetic operations in quantized precision rather than full precision. 
Additionally, uniform quantization may not be optimal in cases where the weight distribution is non-uniform, as often observed in LLMs.
Contrarily, non-uniform quantization offers a solution to these challenges. As studied in SqueezeLLM~\cite{kim2023squeezellm}, this approach allocates quantization bins non-uniformly, allowing for more flexibility and potentially better performance, especially when dealing with non-uniform weight distributions.
}



{
Compared with structured pruning, quantization requires specified hardware to realize the realistic advantage of low-bit precision to reduce the memory cost and inference speedup. For the LLM, due to lack of training data or the computing resources, structured pruning is typically difficult to effectively recover lost knowledge under high compression ratio. However, quantization could typically preserve the performance of LLM effectivelly. Therefore, quantization at the present is more popular and mature used in the LLM compression. It is in sharp contrast to the middle-small model size scenarios, where structured pruning and quantization are both commonly (jointly) used~\cite{yang2020automatic,chen2021only}. 
}
 
\subsection{Low-Rank Decomposition}

{
The weight matrices in a DNN are often low-rank, indicating redundancy in model weights~\cite{sainath2013low,zhu2021geometric,pmlr-v162-zhou22c}. Thus, a natural idea is to factorize the weight matrices into two or more smaller matrices to save parameters. In LLMs, the weight matrices exist in linear layers including self-attention layers and MLP layers, and the embedding layers. There exists studies to factorize these weight matrices for saving parameter quantity and accelerate inference.
}

{
\paratitle{Decomposition on Linear Layer.}  Multi-linear attention~\cite{ma2019tensorized} uses block-term tensor (BTT) decomposition~\cite{de2008decompositions} to factorize multi-head attention. Singular value decomposition (SVD)~\cite{de2008decompositions} is also commonly used and typically performed with a two-stage manner. The first stage is to establish the decomposition followed by a second stage to fine-tune the low-rank weights via knowledge distillation~\cite{noach2020compressing}.  Besides, as an alternative to BTT and SVD, Kronecker decomposition retains the rank of the matrix and has shown improvement during compressing BERT and GPT-2~\cite{tahaei2021kroneckerbert,edalati2021kronecker}.
}

{
\paratitle{Decomposition on Embedding Layer.} ALBERT~\cite{lan2019albert} uses factorization for the embedding layer, which is one of the largest consumers of model parameters. Since the power of Transformer mainly comes from its contextual learning ability, the parameters in the token embedding layer are not efficient. It intuitively makes sense to reduce them by factorizing the embedding matrix. Self-Attentive Factorized embeddings (SAFE)~\cite{reid2021subformer} studied ways to share weights in transformers by adding a small self-attention layer on the basis of linear projection  to achieve better performance than the alternatives. LightFormer~\cite{lv2023lightformer} more effectively utilizes the parameter knowledge of the well-trained Transformer, and accelerates the convergence of the model factorization on the embedding layers.
}

\section{Conclusion}\label{sec:conclusion}

{
In conclusion, the evolution of Large Language Models (LLMs) marks a significant milestone in the field of artificial general intelligence, bringing transformative changes across various domains. However, the rapid expansion of these models brings forth substantial challenges in terms of computational demands and memory requirements, creating hurdles for both academic research and practical deployment. This survey provided a comprehensive overview of the algorithmic innovations aimed at enhancing the efficiency of LLMs, capturing research developments mostly up to September 2023. Moving beyond the scope of the existing surveys that often focus on isolating aspects such as training or model compression, this survey delved into the multiple dimensions of efficiency that are crucial for the holistic algorithmic development of LLMs. It has spanned a broad array of efficiency-related topics including scaling laws, data utilization, architectural designs, as well as training, tuning, and inference strategies. The insights and analyses presented here aim to serve as valuable summarization for both researchers and practitioners in the field. By laying a solid foundation of current knowledge and approaches, this paper sets the stage for future breakthroughs and continued innovation in the crucial research area of LLM efficiency.
}

\bibliographystyle{ACM-Reference-Format}
\bibliography{main}










\end{document}